\documentclass[twoside]{article}

\usepackage[accepted]{aistats2015}
\usepackage{flushend}
\usepackage{graphicx} 
\usepackage{subfigure}
\usepackage{algorithm}
\usepackage{algorithmic}
\usepackage{hyperref}

\usepackage{amssymb,amsthm}
\usepackage{color}
\usepackage{graphicx}
\usepackage{wrapfig}
\usepackage{amsmath,epsfig,subfigure}
\usepackage[authoryear,colon, sort&compress]{natbib}

\usepackage{enumerate}
\usepackage{enumitem}

\newcommand{\beq}{\vspace{0mm}\begin{equation}}
\newcommand{\eeq}{\vspace{0mm}\end{equation}}
\newcommand{\beqs}{\vspace{0mm}\begin{eqnarray}}
\newcommand{\eeqs}{\vspace{0mm}\end{eqnarray}}
\newcommand{\barr}{\begin{array}}
\newcommand{\earr}{\end{array}}

\newcommand{\Bmat}[0]{{{\bf B}}}

\newcommand{\cdotv}{\boldsymbol{\cdot}}

\newcommand{\phiv}{\boldsymbol{\phi}}

\newcommand{\E}{\mathbb{E}}
\newtheorem{thm}{Theorem} 

\newtheorem{lem}[thm]{Lemma}

\begin{document}

%
\runningtitle{Infinite
Edge Partition Models}

%
\runningauthor{Mingyuan Zhou}

\twocolumn[

\aistatstitle{
Infinite
Edge Partition Models for Overlapping\\ Community Detection and Link Prediction 
} 

\aistatsauthor{Mingyuan Zhou}

\aistatsaddress{
McCombs School of Business, 
The University of Texas at Austin, Austin, TX 78712, USA} ]

\begin{abstract}

A hierarchical gamma process infinite edge partition model is proposed to factorize the binary adjacency matrix of an unweighted undirected relational network under a  Bernoulli-Poisson link. The model describes  both homophily and stochastic equivalence, and is scalable to  big sparse networks by focusing its computation on  pairs of linked nodes. It can not only discover overlapping communities and inter-community interactions, but also predict missing edges. A simplified version omitting inter-community interactions is also provided and we reveal its  interesting connections to existing models. The number of communities is automatically inferred in a nonparametric Bayesian manner, and efficient  inference via Gibbs sampling is derived using novel data augmentation techniques. Experimental results on four real networks demonstrate the models' scalability and state-of-the-art performance.

\end{abstract}

\section{INTRODUCTION}

Community detection and link prediction are two important problems in network analysis. 
A vast number of community detection algorithms  based on various useful heuristics, such as modularity maximization  \citep{newman2004finding} and clique percolation \citep{palla2005uncovering}, have been proposed. See \citet{fortunato2010community} for a comprehensive review. These algorithms, however,  are not based on generative models and hence usually cannot   be used to generate networks and predict missing edges (links). Moreover, how to set the number of communities is a critical issue that has not been well addressed by them. 
In this paper, we will fit 
 unweighted undirected relational networks using nonparametric Bayesian generative models, which can be used to simulate random networks, detect latent overlapping communities and community-community interactions, and predict missing edges, with the number of communities automatically inferred from the data.  

For a relational network, a community can be considered as a subset of nodes (vertices) 
that are densely connected to each other but sparsely to the others, such as those in a social network,  or it can be considered as a subset of nodes that are sparsely connected to each other but densely connected to the nodes belonging to another community, such as those in a network consisting of carnivores and herbivores: tigers and bears both hunt deers but  rarely prey on each other. The former phenomenon is usually described as assortativity or homophily, while the latter is known  as dissortativity or stochastic equivalence \citep{hoff2008modeling}. As a relational network may exhibit both homophily and stochastic equivalence, an algorithm capable of modeling  both phenomena would usually be preferred if no prior information on assortativity is available. If analyzing assortative networks with dense intra-community connections is the main goal, then one may consider an assortative algorithm that models homophily but not necessarily stochastic equivalence.

The stochastic blockmodel (SBM) is a popular  latent class model  to detect latent communities \citep{holland1983stochastic,nowicki2001estimation}. 
It partitions the nodes into disjoint communities,  and models the probability for an edge to exist  between two nodes solely based on which two communities that they belong to. 
It is simple and scalable, and models both homophily and stochastic equivalence. In addition, the infinite relational model, a nonparametric Bayesian extension of the SBM based on the Chinese restaurant process \citep{aldous}, allows the number of communities to be automatically inferred from the data
\citep{IRM_Kemp}.  
Despite these attractive properties, the SBM is restrictive in that communities are not allowed to overlap. In practice, however, it is common for a node to belong to multiple communities, motivating the development of more advanced latent class models, such as 
 the mixed-membership stochastic blockmodel (MMSB) of \citet{MMSB} and its various extensions \citep{gopalan2012scalable,
kim2013efficient}.  
The MMSB generalizes the SBM to allow a node to participate in multiple communities, yet
since it has to infer two community indicators for each pair of nodes, regardless of whether an edge exists in that pair, its computation grows quadratically  
as a function of the number of nodes $N$. Moreover, 
the number of communities in the MMSB is a model parameter that needs to be carefully selected. 

In this paper, instead of clustering nodes, as in the SBM, or clustering all possible edges, as in the MMSB, we propose the edge partition model (EPM) to partition only the observed edges, which readily leads to the partition of nodes: 
if the edges linked to a node are partitioned into multiple communities, then the node is naturally affiliated with all these communities, and could be hard assigned to a single community that has the strongest presence in its edges. In contrast to the SBM, the EPM allows  communities to overlap; and in contrast to the MMSB that spends  $O(N^2)$ computation  clustering all possible edges, the EPM spends  $O(\bar{d} N)$ computation  partitioning   only observed edges,  where $\bar{d}$ is the average degree (number of edges) per node, leading to notable computational savings as $\bar{d}$ is often much smaller than $N$ in a big sparse network commonly observed in practice.  

To support a potentially infinite number of communities and to model both homophily and stochastic  equivalence in an unweighted undirected  relational network, we propose a hierarchical gamma process (HGP)  EPM, which links each observed edge to a latent count using a  Bernoulli-Poisson link, and then factorizes the latent $N\times N$ random count matrix. 
The HGP supports the  EPM to have  an infinite dimensional feature vector for each node to describe its affiliations with communities,  and an infinite dimensional square  rate matrix, whose diagonal and off-diagonal elements  describe the intra- and inter- community interactions, respectively. We also propose a gamma process EPM  as a simplified version of the HGP-EPM, which omits inter-community interactions to gain simpler inference and faster computation at the expense of  reduced ability to model stochastic  equivalence.

Conceptually, our idea of directly partitioning edges  and implicitly partitioning nodes into communities is related to the one in \citet{ahn2010link} and \citet{evans2009line}.  In terms of construction, our EPMs are  related to the Poisson factor models of \citet{NewmanCommunityModel} and the Eigenmodel of \citet{hoff2008modeling}. In terms of supporting an infinite number of features, 
our EPMs are related to the models in  \citet{miller2009nonparametric} and 
  \citet{morup2011infinite} that use the Indian buffet process of \citet{IBP} to support an infinite binary feature matrix. 
The proposed models depart from  existing ones with several distinctions: 1) a Bernoulli-Poisson link connects each edge to a latent count that is further partitioned;  2) a hierarchical gamma process is constructed to support an infinite number of communities and an infinite-dimensional square matrix to describe community-community interactions;  3) two nonparametric Bayesian EPMs are constructed to factorize the $N\times N$ binary adjacency matrix under the Bernoulli-Poisson link, supporting a nonnegative feature matrix with an unbounded number of columns, and at the same time assign each edge and hence each node to one or multiple latent communities; and 4) efficient and scalable Bayesian inference via Gibbs sampling 
is provided.

\vspace{-1mm}
\section{FACTOR ANALYSIS AND BERNOULLI-POISSON LINK}
\vspace{-2mm}

Our basic idea is to factorize the BINARY network adjacency matrix  using tools developed for COUNT data analysis, and to discover overlapping communities and their interactions by examining how the latent count for each edge is  partitioned. This section will primarily discuss individual model components and their properties, with hierarchical Bayesian models presented later. 

\subsection{Poisson Factor Analysis}\label{sec:PFM2}
We propose a Poisson factor model  for a weighted undirected $N\times N$ relational network  as
\beqs
&m_{ij}\sim \mbox{Po}\left( \sum_{k_1=1}^K \sum_{k_2=1}^K \phi_{ik_1}\lambda_{k_1 k_2}\phi_{jk_2}\right),\label{eq:PFM2}
\eeqs
where $m_{ij}\equiv m_{ji}$ is the integer-valued weight (observed or latent) that links nodes  $i$ and $j$, $(\phi_{i1},\ldots,\phi_{iK})$ is the positive feature vector for node $i$, $\lambda_{k_1k_2}\equiv \lambda_{k_2k_1}$ is a positive rate, and the symbol $\equiv$ denotes ``equal by definition.'' This model is conceptually simple: 
with $\phi_{ik_1}$ measuring how strongly node~$i$ is affiliated with  community $k_1$  and  $\lambda_{k_1k_2}$ measuring  how strongly communities $k_1$ and $k_2$ interact with each other, the product $\phi_{ik_1}\lambda_{k_1 k_2}\phi_{jk_2}$ measures how strongly nodes~$i$ and $j$ are connected 
due to their affiliations with communities $k_1$ and $k_2$, respectively, and 
a weighted combination of all intra-community weights $\{\lambda_{kk}\}_{1\le k \le K}$ and inter-community ones $\{\lambda_{k_1k_2}\}_{1\le k_1\neq k_2 \le K}$ is the expected value of $m_{ij}$. 

The factor model in (\ref{eq:PFM2}) makes intuitive sense.  For example, suppose persons $i$ and $j$ are  both residents of City Avatar and   active members of the Avatar anglers Meetup 
 group that organizes fishing trips regularly. In addition, persons $i$ is an active member of the Avatar artificial intelligence (AI) Meetup group while person $j$ is an active member of the Avatar statistics Meetup group. 
 Denoting $m_{ij}$ as the number of times  that $i$ and $j$ 
attend the same group meeting in 2015, 
then due to their strong affiliations with the anglers Meetup group, $m_{ij}$ would have a large expected value, which is likely to be further increased if the AI and statistics Meetup groups hold joint events regularly. 

To model an assortative relational network exhibiting homophily but not necessarily stochastic equivalence, we may omit the inter-community interactions by letting $\lambda_{k_1k_2}\equiv 0$ for $k_1\neq k_2$ and  simplify (\ref{eq:PFM2}) as
\beqs
&m_{ij}\sim\mbox{Po}\left(\sum_{k=1}^K r_k \phi_{ik}\phi_{jk}\right),\label{eq:PFM1}
\eeqs
where $r_k\equiv \lambda_{kk}$ 
indicates the prevalence of community~$k$, and two nodes with similar latent features are encouraged to be linked by an edge with a large weight. 

 We note \citet{NewmanCommunityModel} had examined a model related to (\ref{eq:PFM1}) and briefly mentioned a model related to 
(\ref{eq:PFM2}). However, they used a heuristic approach to model binary data under the Poisson distribution, did not provide a  principled way to set the number of communities $K$, and had to create possibly nonexistent self-edges  in order to derive tractable expectation-maximization (EM) inference. This paper will address all these issues rigorously, in a nonparametric Bayesian manner, and carefully examine the models in both (\ref{eq:PFM1}) and (\ref{eq:PFM2}) and provide efficient Bayesian inference.

\vspace{-1mm}
\subsection{Bernoulli-Poisson Link }\label{sec:PoBer}
\vspace{-1mm}

To use the Poisson factor models in (\ref{eq:PFM2}) and (\ref{eq:PFM1}) for an unweighted network with a binary adjacency matrix, we introduce a Bernoulli-Poisson (BerPo) link function that thresholds a random count at one to obtain a random variable in $\{0,1\}$ as
\beq
b =\mathbf{1}(m\ge 1),~ m\sim\mbox{Po}(\lambda),\label{eq:BerPo}
\eeq
where $b=1$ if $m\ge 1$ and $b=0$ if $m=0$.
The intuition is that two nodes are connected if they interact at least once. The mathematical motivation is after transforming a binary-modeling problem  into a count-modeling one, one is readily equipped with a rich set of statistical tools developed for count data analysis using the Poisson and negative binomial distributions. 


If $m$ is marginalized out from (\ref{eq:BerPo}), then given $\lambda$, one obtains a  Bernoulli random variable as
\beq
b\sim\mbox{Ber}\left(1-e^{-\lambda}\right).\notag 
\eeq 
The conditional posterior of $m$ can be expressed as
$$(m|b,\lambda)\sim b\cdotv \mbox{Po}_{+}(\lambda),$$ where $x\sim\mbox{Po}_{+}(\lambda)$ follows a truncated Poisson distribution, with 
$P(x=k) =(1-e^{-\lambda})^{-1} {\lambda^k e^{-\lambda}}/{k!}$ for $k\in\{1,2,\ldots\}$.
Thus if $b=0$, then $m= 0$ almost surely (a.s.), and if $b=1$, then 
$m\sim\mbox{Po}_{+}(\lambda)$, which can be simulated with rejection sampling: if $\lambda\ge 1$, we draw $m\sim\mbox{Po}(\lambda)$ till $m\ge 1$; and if $\lambda<1$, we draw both $n\sim\mbox{Po}(\lambda)$ and $u\sim\mbox{Unif}(0,1)$ till $u<1/(n+1)$, and then let $m=n+1$. The acceptance rate is $1-e^{-\lambda}$ if $\lambda\ge1$ and ${\lambda^{-1}}(1-e^{-\lambda})$ if $\lambda<1$, and reaches its minimum,  63.2\%, when $\lambda=1$.

The BerPo link shares some similarities with the probit link that thresholds a normal random variable at zero, and the logit link that lets $b\sim\mbox{Ber}[e^{x}/(1+e^{x})]$. 
We advocate the BerPo link as an alternative to the probit and logit links since if $b=0$, then $m= 0$ a.s., which could lead to significant computational savings if a considerable proportion of the data are equal to zero. In addition,  the additive property of the Poisson allows us to model the link strength  between any two nodes by aggregating the contributions of all possible intra- and inter- community interactions, 
and the conjugacy between the Poisson and gamma distributions makes it convenient to construct hierarchical Bayesian models amenable to posterior simulation.

\subsection{Overlapping Community Structures}
Note that (\ref{eq:PFM2}) can be augmented as
\beq\small
m_{ij}=\hspace{-1mm}\sum_{k_1}\sum_{k_2} m_{ik_1k_2j}, ~m_{ik_1k_2j}\sim \mbox{Po}\left( \phi_{ik_1}\lambda_{k_1 k_2}\phi_{jk_2}\right).\label{eq:PFM2_augmented}\hspace{-1mm}\vspace{-1mm}
\eeq\normalsize
where $m_{ik_1k_2j}$ represents how often nodes~$i$ and $j$ interact 
due to their affiliations with communities $k_1$ and $k_2$, respectively. We may consider that the model is partitioning the count $m_{ij}$ into $\{m_{ik_1k_2j}\}_{1\le k_1,k_2\le K}$, and hence
we call the Poisson factor model in (\ref{eq:PFM2})  together with the BerPo link in (\ref{eq:BerPo}) as an edge partition model (EPM), in which each edge is partitioned according to all possible $K^2$ community-community interactions, and
how strongly node $i$ is affiliated with  community $k$ can be measured with $\phi_{ik}\omega_{ik}$, where 
\beqs
&\omega_{ik}:=
\sum_{j\neq i}\sum_{k'} \phi_{jk'}\lambda_{k k'}\label{eq:omegaik}\vspace{-1mm}
\eeqs
represents how strongly node $i$ would interact with all the other nodes through its affiliation with community~$k$. 
We further introduce the latent count
\beqs
&m_{i k \cdotv \cdotv} := \sum_{j>i}\sum_{k_2} m_{i k k_2 j} + \sum_{j<i}\sum_{k_1} m_{j k_1 k i},\label{eq:midotkdot}\vspace{-1mm}
\eeqs
to represent how often node $i$ is connected to the other nodes due to its affiliation with community $k$. 
We can then assign node $i$ to multiple communities in $\{k:m_{i k \cdotv \cdotv} \ge 1\}$, or  (hard) assign it to a single community using 
either 
 $\underset{k}{\operatorname{argmax}} ( \phi_{ik}\omega_{ik})\notag$ or $\underset{k}{\operatorname{argmax}} ( m_{i k \cdotv \cdotv} )$.
Similar analysis applies to  a simpler EPM built on (\ref{eq:PFM1}).
By hard assigning each node to a single community and ordering the nodes from the same community to be adjacent to each other, we expect the ordered  adjacency matrix to exhibit a block structure, where the blocks along and off the diagonal represent the intra- and inter- community connections, respectively.

\vspace{-1mm}
\subsection{Scalability for Big Sparse Networks}
\vspace{-1mm}
We are motivated to construct the EPMs because they not only  allow each edge and hence each node to 
participate in multiple communities, 
but also readily scale to a   big sparse network whose average degree per node is much smaller than $N$. A key observation for scalable computation is that (\ref{eq:PFM2}) can be augmented as (\ref{eq:PFM2_augmented}),
where $m_{ik_1k_2j}= 0$  a.s. for any $k_1$ and $k_2$ 
if no edge exists between nodes $i$  and  $j$  (i.e., $m_{ij}=0$). 
On a sparse network, where the  edges 
constitute  only a small portion of all possible $N^2$ edges,  this property makes the EPMs computationally appealing. By contrast, conceptually related models, including the MMSB of \citet{MMSB}, Eigenmodel of \citet{hoff2008modeling} and latent feature relational model of \citet{miller2009nonparametric},   spend computation indiscriminately on all pairs  of nodes $(i,j)$ no matter whether an edge exists between nodes $i$ and $j$, and hence they  have $O(N^2) $ computation and do not scale well as $N$ increases. 

\section{EDGE PARTITION MODELS}
\vspace{-1mm}
\subsection{Hierarchical Gamma Process}
\vspace{-1mm}
The 
EPM takes a weighted combination of all possible intra- and inter- community weights to explain each pair of node, however, the number of communities $K$ is still a model parameter that needs to be set appropriately. 
To  allow $K$ to be inferred from the data and potentially grow to infinity, we need  to introduce a stochastic process that can generate a countably infinite number of atoms $\{\phiv_{k}\}_{1,\infty}$, where $\phiv_{k}=(\phi_{1k},\ldots,\phi_{Nk})^T$ measures how strongly the $N$ nodes are affiliated with  community $k$, and an  infinite dimensional square matrix $\{\lambda_{k_1 k_2}\}_{1\le k_1, k_2\le\infty}$, where $\lambda_{k_2k_1}=\lambda_{k_1k_2}$ measures how strongly communities $k_1$ and $k_2$ interact with each other. Moreover, we need to ensure $\sum_{k_1=1}^\infty \sum_{k_2=1}^\infty \lambda_{k_1k_2}$ 
to be finite a.s. and we may wish to impose some structural regularization on  the infinite square matrix. 

To satisfy all these needs, 
we 
first define
\beq \label{eq:GammaP}
G\sim\Gamma\mbox{P}(G_0,1/c_0)
\eeq
as a gamma process on a product space $\mathbb{R}^+\times \Omega$, where $\mathbb{R}^+=\{x:x>0\}$,  $\Omega$ is a complete separable metric space, $1/c_0$ is a positive scale parameter, and $G_0$ is a finite and continuous base measure, such that $G(A)\sim\mbox{Gam}(G_0(A),1/c_0)$ for each Borel set $A\subset\Omega$ \citep{ferguson73,PoissonP}. The L\'evy measure of the gamma process can be expressed as $\nu(drd\phiv)=r^{-1}e^{-c_0 r}drG_0(d\phiv)$,  
and 
a draw from the gamma process, consisting of countably infinite atoms, can be expressed as $G=\sum_{k=1}^\infty r_k\delta_{\phiv_k}$, where $\phiv_k\stackrel{iid}{\sim}g_0$, $G_0=\gamma_0 g_0$, $g_0(d\phiv)=  G_0(d\phiv)/\gamma_0$ is the base distribution, and $\gamma_0=G_0(\Omega)$ is the mass parameter. A gamma process based model has an inherent shrinkage mechanism, as in the prior the number of atoms with $r_k$ greater than  $\varepsilon\in\mathbb{R}^+$ follows $\mbox{Po}(\gamma_0\int_{\varepsilon}^\infty r^{-1}e^{-cr}dr)$, whose Poisson rate decreases as $\varepsilon$ increases.

Given $G=\sum_{k=1}^\infty r_k\delta_{\phiv_k}$, we further define 
a relational gamma process ($\mbox{r}\Gamma\mbox{P}$) as
\beq\label{eq:rGammaP}
\Lambda|G \sim\mbox{r}\Gamma\mbox{P}(G,\xi,1/\beta),
\eeq
a draw from which is defined as
\beqs
&\Lambda = \sum_{k_1=1}^\infty \sum_{k_2=1}^\infty \lambda_{k_1k_2}\delta_{(\phiv_{k_1},\,\phiv_{k_2})}, \notag\vspace{-1mm} 
\eeqs
where $\xi$ and $\beta$ are both in $\mathbb{R}^+$, $\lambda_{k_2k_1}\equiv \lambda_{k_1k_2}$, and
\begin{align}
&\lambda_{k_1 k_2}\sim\begin{cases} \vspace{0.15cm} \mbox{Gam}(\xi r_{k_1},1/\beta)\
, & \mbox{if }  k_2=k_1, \\
\mbox{Gam}( r_{k_1} r_{k_2},1/\beta), 
& \mbox{if  }  k_2 > k_1.\notag
\end{cases}
\end{align}
Given a relational  gamma process draw $\Lambda$, we 
generate a binary adjacency matrix $\Bmat\in\{0,1\}^{N\times N}$ as 
\beq\label{eq:Mmat}
\Bmat|\Lambda\sim\mbox{Ber}\left[1-\prod_{k_1=1}^\infty\prod_{k_2=1}^\infty \exp\left(-\phiv_{k_1}\lambda_{k_1k_2}\phiv_{k_2}^T\right)\right].\!\!
\eeq
Equations (\ref{eq:Mmat}),  (\ref{eq:rGammaP}) and (\ref{eq:GammaP}) constitute an HGP-EPM that supports countably infinite atoms and a countably  infinite square matrix, the total sum of whose elements has a finite expectation, as shown  in the following Lemma, with proof provided in the Appendix. 

\begin{lem}\label{lem:E_lambda}
 The expectation  of $\sum_{k_1=1}^\infty\sum_{k_2=1}^\infty \lambda_{k_1 k_2}$ 
 is finite and 
 can be expressed as 
\beq
\E\bigg[\sum_{k_1}\sum_{k_2} \lambda_{k_1 k_2}\bigg]  = \frac{\xi}{c_0 \beta}\notag\gamma_0 + \frac{1}{c_0^2 \beta}\gamma_0^2\label{eq:E_lambda}.\, \vspace{-3.3mm}
\eeq
\end{lem}

The usual scenario to consider an HGP construction is when one models grouped data and wishes to share statistical strengths across groups. For example, the gamma-negative binomial process of \citet{MingyuanNIPS2012}, related to the hierarchical Dirichlet process of \citet{hdp}, is considered for topic modeling, where each document is associated with a gamma process, and these gamma processes are coupled by sharing a lower-level ($i.e.$, further from the data) gamma process as their atomic base measure. The proposed HGP is distinct 
 in that the product of the weights of any two atoms of the lower-level gamma process is used to parameterize the shape parameter of a gamma random variable higher in the hierarchy. 

The proposed HGP also helps express our prior belief that an atom with a small weight tends to represent a small community, which also tends to interact with the others 
less frequently. Note that if we let $\lambda_{k k}\sim \mbox{Gam}(r_k^2,1/\beta)$, then 
 the expectation of the matrix $\{\lambda_{k_1k_2}\}_{1\le k_1,k_2\le\infty}$ given $\{r_k\}_{1,\infty}$ has a rank of one. We use $\xi r_k$ instead of  $ r_k^2$ as the shape parameter of  $\lambda_{k k}$ to allow $r_k$ to be inferred with Gibbs sampling and to prevent  overly shrinking $\lambda_{kk}$ 
for small communities. 
Note that \citet{palla2014reversible} proposed a reversible infinite hidden Markov model using a related HGP infinite square rate matrix, the normalization of whose each row represents a state transition probability  vector. Our HGP serves a distinct modeling purpose; 
no normalization is required for the  infinite square rate matrix, and our model allows exploiting unique data augmentation techniques to infer both $\lambda_{k_1k_2}$ and $r_k$ with closed-form Gibbs sampling update equations, as discussed in Section \ref{sec:MCMC} and the Appendix.

\vspace{-1.5mm}
\subsection{
Hierarchical Gamma Process EPM 
}
\vspace{-1mm}

We choose the base distribution of the gamma process  $G\sim\Gamma\mbox{P}(G_0,1/c_0)$  as $g_0(\phiv) = \prod_{i=1}^N \mbox{Gam}(a_i,1/c_i)$. 
For implementation convenience, we consider a discrete base measure as $G_0=\sum_{k=1}^K \frac{\gamma_0}{K} \delta_{\phiv_k}$, where $K$ is a truncation level that is set large enough to ensure a good approximation to the truly infinite model. 
We express the (truncated)  HGP-EPM  as

\vspace{-6mm}
\begin{align}
&b_{ij} = \mathbf{1}(m_{ij}\ge 1),~m_{ij}=\sum_{k_1=1}^K \sum_{k_2=1}^K m_{ik_1k_2j},\notag\\
&m_{ik_1k_2j} \sim\mbox{Po}\left( \phi_{ik_1}\lambda_{k_1 k_2}\phi_{jk_2}\right),\notag\\
&\phi_{ik}\sim\mbox{Gam}(a_i,1/c_{i}),~a_i\sim\mbox{Gam}(e_0,1/f_0),\notag\\
&\lambda_{k_1 k_2}\sim\begin{cases} \vspace{0.15cm} \mbox{Gam}( \xi  r_{k_1},1/\beta),\, & \mbox{if }  k_2=k_1, \\
\mbox{Gam}( r_{k_1} r_{k_2},1/\beta), 
& \mbox{if  }  k_2> k_1,
\end{cases}\notag\\
&r_k\sim\mbox{Gam}(\gamma_0/K,1/c_0)
\label{eq:BPFM2},
\end{align}
\vspace{-6mm}

where $\lambda_{k_2 k_1}\equiv \lambda_{k_1 k_2}$  and conjugate gamma priors are  imposed on  $\gamma_0$, $\xi$, $c_0$, $c_i$ and $\beta$.
Note that marginalizing out  both $m_{ij}$ and $m_{ik_1k_2j}$ from (\ref{eq:BPFM2}) leads to
\beq
b_{ij}\sim\mbox{Ber}\bigg[1 - \prod_{k_1=1}^K\prod_{k_2=1}^K \exp(- \phi_{ik_1}\lambda_{k_1 k_2}\phi_{jk_2})\bigg].
\label{eq:BerPoLink}
\eeq
A noticeable  advantage of  the augmented representation in (\ref{eq:BPFM2}) over  (\ref{eq:BerPoLink}) is that  (\ref{eq:BPFM2}) is amenable to posterior simulation, as discussed in Section \ref{sec:MCMC}. 

Note that similar to  \citet{hoff2008modeling} and \citet{lloyd2012random}, we assume that the nodes are exchangeable and hence the discussions of \citet{Hoover} and \citet{aldous} on exchangeability also apply to our EPMs. 
 
\vspace{-1mm}
\subsection{Gamma Process EPM}\label{sec:GPEPMandAGM}
\vspace{-1mm}
If we omit inter-community interactions by letting $\lambda_{k_1k_2}\equiv 0$ for $k_1\neq k_2$ and $\lambda_{kk}\equiv r_k$,
then the HGP-EPM reduces to a gamma process EPM (GP-EPM), which is likely to well fit assortative networks but not necessarily disassortative ones. 
We  notice an interesting connection to  the community-affiliation graph model (AGM) of \citet{yang2012community,YangAGM2014}: the GP-EPM generates 
an edge 
with probability 
\beqs
&\!\!P(b_{ij}\!=\!1) = 1\!- \prod_k\left\{1-[1\!-\!\exp(-r_k\phi_{ik}\phi_{jk})]\right\}; \,\,\,\,\, \,\,\label{eq:IEPM} 
\eeqs
if we define $p_k=1-e^{-r_k}$ and further impose the restriction that  $\phi_{ik}\in\{0,1\}$, then (\ref{eq:IEPM}) reduces to
\beqs
&P(b_{ij}=1) = 1- \prod_{k\in C_{ij}}\left(1-p_k\right),\label{eq:AGM}\vspace{-1.5mm}
\eeqs
where $C_{ij}=\{k: \phi_{ik}=1 \text{~and~}  \phi_{jk}=1\}\subset\{1,\ldots,K\}$ is a set of communities that nodes $i$ and $j$ share; note that (\ref{eq:AGM}) 
is almost the same as the AGM  of \citet{yang2012community,YangAGM2014}. 
In fact, one may consider  the GP-EPM with $b_{ij}\sim \mbox{Ber}[1 - e^{-\epsilon}\prod_k\exp(-r_k\phi_{ik}\phi_{jk})],$ where $\epsilon\in\mathbb{R}^+$ and $\phi_{ik}\in\{0,1\}$,  as a nonparametric Bayesian   AGM. 
Similarly, we also notice that (\ref{eq:BerPoLink}) of the HGP-EPM is related to the model 
of \citet{morup2011infinite} if we restrict $\phi_{ik}\in\{0,1\}$. 

 \citet{yang2012community,YangAGM2014} argue that all previous community detection methods, including clique percolation and MMSB, would fail to detect communities with dense overlaps, because they all had a hidden assumption that a community's overlapping parts are less densely connected than its non-overlapping ones. The same as the AGM,  both the GP-EPM and HGP-EPM  do not make such a restrictive assumption, and they both 
allow  overlaps of communities to be denser than communities themselves; Beyond the AGM, we do not restrict $\phi_{ik}$ to be either zero or one, and our generative models are built under a rigorous nonparametric Bayesian  framework with  efficient Bayesian inference, as presented below. 

\vspace{-2mm}
\subsection{MCMC Inference}\label{sec:MCMC}
\vspace{-2mm}
In this paper, we consider an unweighted  undirected network, where $b_{ji}\equiv b_{ij}$  and  self-links $b_{ii}$ are not defined. Thus we only consider $b_{ij}$ for $j>i$ in (\ref{eq:BPFM2}). Let $m_{i k \cdotv \cdotv}$ be defined as in (\ref{eq:midotkdot}) and 
$m_{\cdotv k_1 k_2 \cdotv}$ as 
\beqs
&m_{\cdotv k_1 k_2 \cdotv}:= 2^{-\delta_{k_1k_2}}\sum_{i}\sum_{j>i}( m_{i k_1 k_2 j} +  m_{i k_2 k_1 j}),\notag\vspace{-1mm}
\eeqs
where $\delta_{k_1k_2}=1$ if $k_1=k_2$ and $\delta_{k_1k_2}=0$ otherwise. 
Using (\ref{eq:omegaik}) and the Poisson additive property, we have
\begin{align}
m_{i k \cdotv \cdotv} & \sim\mbox{Po}( \phi_{ik} \omega_{ik} 
)\label{eq:phi_ik_likelihood},\\
m_{\cdotv k_1 k_2 \cdotv}  &\sim\mbox{Po}\left(\lambda_{k_1k_2}  \theta_{k_1k_2} 
\right )
\label{eq:lambda_likelihood},
\end{align}
where $\theta_{k_1k_2} := 2^{-\delta_{k_1k_2}} \sum_i\sum_{j\neq  i}\phi_{ik_1}\phi_{jk_2}$ represents how strongly the nodes interact through communities $k_1$ and $k_2$.
Marginalizing out $\phi_{ik}$ from (\ref{eq:phi_ik_likelihood}) and $\lambda_{k_1k_2}$ from (\ref{eq:lambda_likelihood}), with $p_{ik}':= \omega_{ik}/(c_i+\omega_{ik})$ and $
\tilde{p}_{k_1 k_2} := \theta_{k_1k_2}/(\beta + \theta_{k_1k_2})$, we have
 \begin{align}
 m_{i k \cdotv \cdotv}  & \sim \mbox{NB}\left( a_i, p_{ik}'\right),
 \label{eq:a_i_likelihood}\\
m_{\cdotv k_1 k_2 \cdotv} & \sim\mbox{NB}\left[ r_{k_1} \xi^{\delta_{k_1k_2}} (r_{k_2})^{1-\delta_{k_1k_2}} ,~\tilde{p}_{k_1 k_2}\right ]\label{eq:r_k_likelihood}.
 \end{align}
 
Using the BerPo link, the gamma-Poisson conjugacy, and the  augment-and-conquer techniques to infer the negative binomial dispersion parameters \citep{MingyuanNIPS2012,NBP2012}, 
we exploit  (\ref{eq:phi_ik_likelihood})-(\ref{eq:r_k_likelihood}) to derive 
closed-form Gibbs sampling update equations for all model parameters except $\gamma_0$, and construct an excellent proposal distribution to sample $\gamma_0$ using an independence chain Metropolis-Hastings algorithm. 
We present in the Appendix the details of MCMC inference for the HGP-EPM, and the hierarchical model and closed-form Gibbs sampling update equations for the GP-EPM. The inference of the nonparametric Bayesian AGM would be almost the same as that of the GP-EPM, with the only difference that  the $(\phi_{ik}|-)$ would be sampled from Bernoulli distributions.

 \vspace{-2mm}
\section{EXPERIMENTAL RESULTS}
\vspace{-2mm}

 For comparison,  we consider the infinite relational model (IRM) of \citet{IRM_Kemp}, the Eigenmodel of \citet{hoff2008modeling}, the infinite latent attribute (ILA) model of  \citet{ILA_ICML2012},  the AGM of  \citet{yang2012community,YangAGM2014}, and our GP- and HGP-EPMs. 
  We use the R package provided for the Eigenmodel. We use the ILA code\footnote{\href{http://mlg.eng.cam.ac.uk/konstantina/ILA/ILA_code(v1).tar.gz}{\tiny http://mlg.eng.cam.ac.uk/konstantina/ILA/ILA$\_$code(v1).tar.gz}} provided for \citet{ILA_ICML2012}, in which it is shown that the ILA outperforms the related  nonparametric latent feature relational model of \citet{miller2009nonparametric}. 
    We implement a nonparametric Bayesian version of the AGM as a special case of the GP-EPM, as discussed in Section \ref{sec:GPEPMandAGM}. 
 Matlab code for the EPMs is available on the author's website.

 For the Eigenmodel,  we find the best $K$ in $\{5,10,25,50\}$.  For the ILA, we use its default parameter setting\footnote{
The default training/testing partition of the ILA code sends self-edges into the testing set; whereas in this paper, we do not intend to predict self-edges and hence we do not allow them to appear in the testing set. 
}. For the IRM, we choose $\mbox{Beta}(0.1,1)$ as the prior for each latent block   and $\mbox{Gam}(0.01,1/0.01)$ as the prior for the Chinese restaurant process concentration parameter; for the nonparametric Bayesian AGM, we let $\phi_{ik}\sim\mbox{Ber}(\pi_i),~\pi_i\sim\mbox{Beta}(0.01,0.01)$ and $\epsilon\sim\mbox{Gam}(0.01,1/0.01)$; these parameters are found to consistently provide good performance.  For our models' hyper-parameters, we choose $e_0=f_0=0.01$ and let $\gamma_0$, $c_i$, $c_0$ and $\beta$ be all drawn from $\mbox{Gam}(1,1)$. 
  
  We consider 3000 MCMC iterations and collect the last 1500 samples, unless otherwise stated. 
 We consider two small-scale benchmark networks, for which we test all algorithms and set the truncation level as $K_{\max}=100$ for our algorithms, and another two   networks with more than 2000 nodes, for which we set  $K_{\max}=256$.

To test a model's ability to predict missing edges of an unweighted undirected relational network,  we randomly\footnote{If removing an edge disconnects a node to all the others, then the edge will be kept in the training set.} hold out 20\% pairs of nodes and use the the remaining 80\% to predict the probability for an edge to exist in each of these held-out pairs. 
Letting $o_{ij}=0$ if $b_{ij}$ is held out and $o_{ij}=1$ otherwise, we only need to slightly modify the inference by only considering $\{b_{ij}:o_{ij}=1\}$  in the likelihood. For example,
$\omega_{ik}$  in (\ref{eq:omegaik}) would be redefined as $\omega_{ik} = \sum_{j:o_{ij}=1}\sum_{k'} \lambda_{k k'}  \phi_{jk'}$.
 We consider exactly the same five random training-testing partitions for all algorithms and report the average area under the curve (AUC) of both the receiver operating characteristic (ROC) and  precision-recall (PR) curves \citep{davis2006relationship}. For link prediction, the AUC-PR is more sensitive to 
  the percentage of true edges among the top ranked ones. 
 Note that in addition to link prediction,   the HGP-EPM, GP-EPM, AGM and IRM all have easily interpretable latent representations that will be used to detect overlapping/disjoint communities. 

 \vspace{-1.5mm}
  \subsection{Protein230 Network}
   \vspace{-1.5mm}

 We first consider the Protein230 dataset of  \citet{butland2005interaction} that describes the  interactions between 230 proteins, with 595 edges. This is a small-scale benchmark network that exhibits both homophily and stochastic equivalence, as shown in \citet{hoff2008modeling} and also tested in \citet{lloyd2012random}. We are able to run 3000 MCMC iterations quickly enough for all algorithms except for the ILA on this network.

 As shown in Tab. \ref{tab:Protein230}, the HGP-EPM has the best overall performance. The Eigenmodel is the second best with $K=10$ and the IRM is the third best. The AGM is not competitive as it restricts its features to be binary. In this and all future tables, we highlight in bold both the best result and the ones that are less than one standard error away from the best. Below we analyze why the HGP-EPM performs the best while  the simpler GP-EPM is not that competitive on this dataset.

 \begin{figure}[!tb]
\begin{center}
\includegraphics[width=63mm]{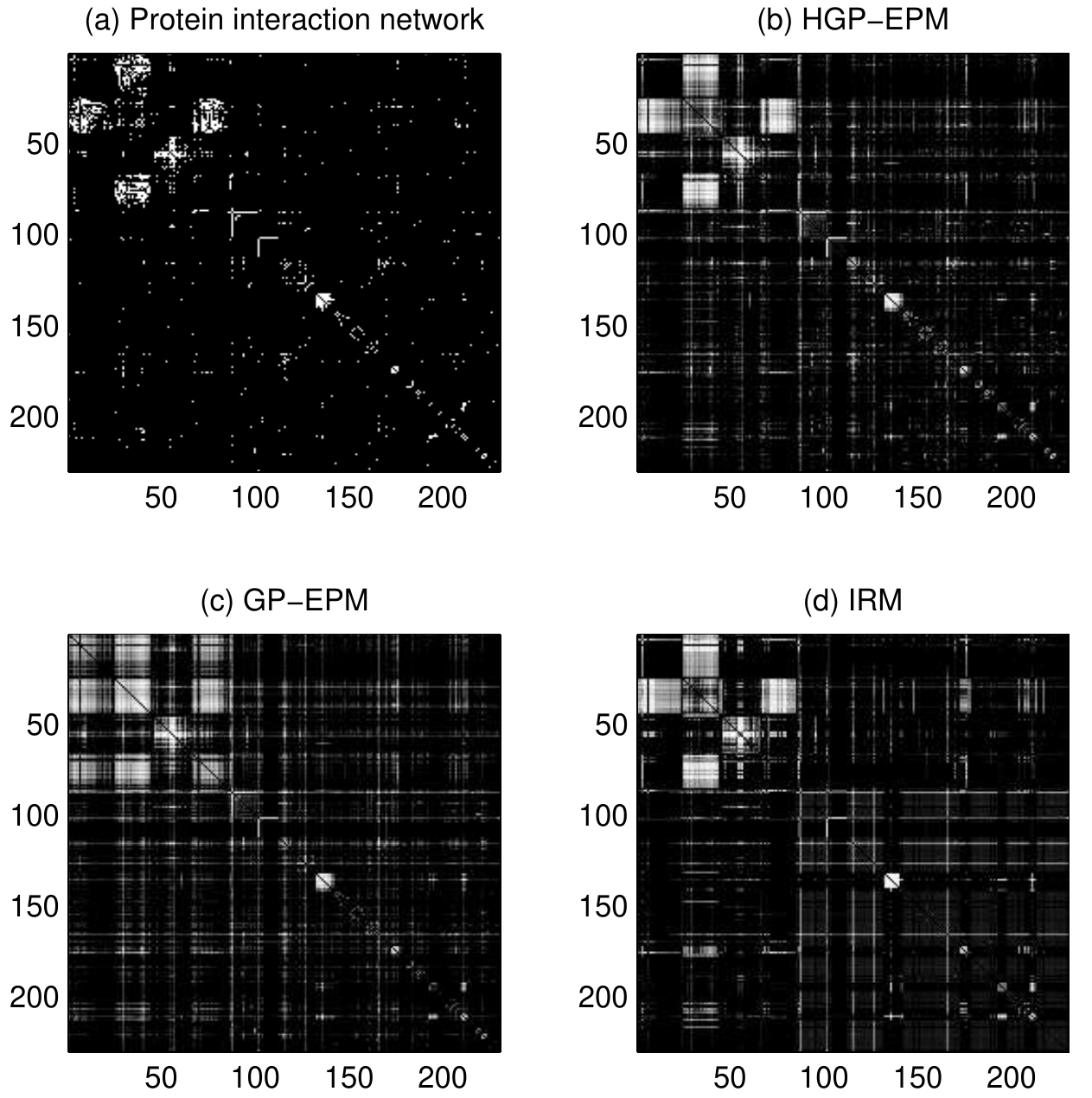}
\end{center}
\vspace{-3.9mm}
\caption{\small\label{fig:protein}
Comparison of three models on estimating the link probabilities for the Protein230  network using 80\% of its node pairs. The nodes are reordered to make a node with a larger index belong to the same or a smaller-size community, where the disjoint community assignments are obtained by analyzing the results of the HGP-EPM. (a) The binary adjacency matrix. 
(b)-(c)  Estimated link probabilities displayed on the log-10 scale from $-2$ to $1$, with a brighter color representing a higher link probability. 
\vspace{-3.8mm}
}
\end{figure}
 
  \begin{table}[!tb] \vspace{-2mm}\small
\caption{\small Comparison of six algorithms on predicting missing edges of the Protein230 network. The Eigenmodel achieves its best performance at $K=10$.
} 
\begin{center}
\begin{tabular}{|c|c|c|c|c|}
\hline
Model & AUC-ROC & AUC-PR\\\hline
IRM &   0.9338 $\pm$ 0.0128 &  0.5026 $\pm$ 0.0676   \\\hline
Eigenmodel &  0.9314 $\pm$ 0.0188 &  \textbf{0.5468} $\pm$  0.0500  \\\hline
ILA &  0.8971 $\pm$ 0.0297 & 0.3693 $\pm$  0.0234  \\\hline
AGM &0.9145 $\pm$ 0.0160 &  0.3339 $\pm$ 0.0359\\\hline
GP-EPM &0.9335 $\pm$ 0.0110 &  0.4011 $\pm$ 0.0452\\\hline
HGP-EPM & \textbf{0.9519} $\pm$ 0.0100 &  \textbf{0.5655} $\pm$ 0.0505 \\\hline
\end{tabular}
\end{center}
\label{tab:Protein230}
\vspace{-5mm}
\end{table}

 As shown in Figs. \ref{fig:protein} (b)-(d), the HGP-EPM captures both homophily and stochastic equivalence by accurately modeling  both diagonal and off-diagonal dense regions of the adjacency matrix; the GP-EPM captures homophily  by accurately modeling diagonal dense regions that represent intra-community interactions, but at the expense of creating nonexistent  blocks in order to fit dense  off-diagonal regions that represent strong inter-community interactions; and the IRM captures these  large dense blocks, but produces  a cartoonish estimation, which overlooks small communities that represent fine details along the diagonal. 
 
\begin{figure}[!tb]
\begin{center}
\includegraphics[width=70mm]{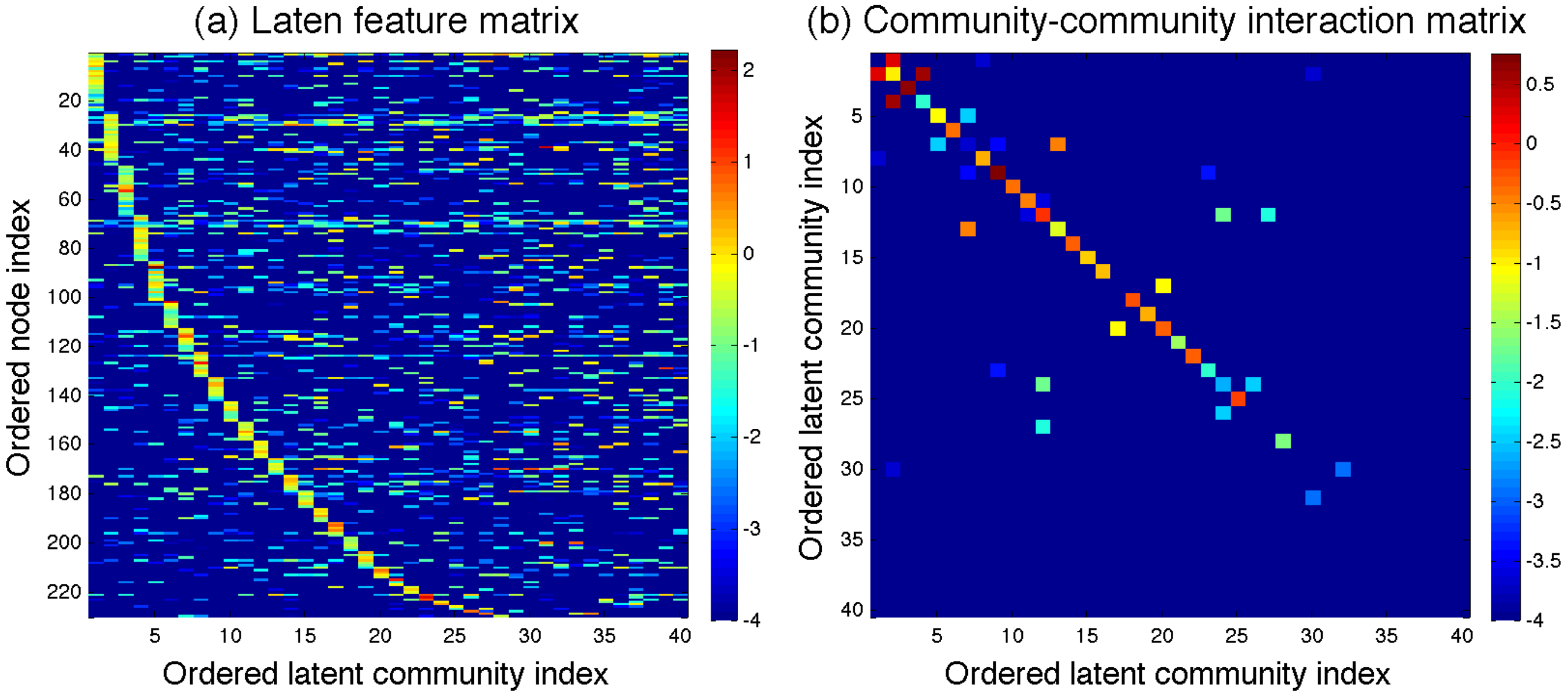}
\end{center}
\vspace{-4.3mm}
\caption{\small\label{fig:protein_matrix}
The inferred latent feature matrix $\{\phiv_k\}$ and community-community interaction rate matrix $\{\lambda_{k_1k_2}\}$ for the HGP-EPM on Protein230.  The nodes are reordered to make a node with a larger index belong to the same or a smaller-size community, and the communities are ordered to make a community with a larger index to have a smaller size. The pixel values are displayed on the log-10 scale. 
\vspace{-4.0mm}
}
\end{figure}

 Fig. \ref{fig:protein_matrix} shows  how the HGP-EPM works. First, each feature vector $\phiv_{k}$ shown in Fig. \ref{fig:protein_matrix} (a) clearly describes how strongly the nodes are affiliated with the community it represents, and each node may have large weights on multiple community. Second, about 30 latent feature vectors are inferred and the remaining ones are essentially drawn from the prior $\prod_i \mbox{Gam}(a_i,1/c_i)$. Third, the inter- and intra-community interaction strengths in  Fig. \ref{fig:protein_matrix} (b) can be matched to the corresponding communities (subsets of nodes) in Figs. \ref{fig:protein} (a) and (b). For example, Fig. \ref{fig:protein_matrix} (a) suggests that the first and second largest communities have  24 and 22  nodes, respectively, and Fig. \ref{fig:protein_matrix} (b) suggests that the first and second communities have sparse and dense intra-community connections, respectively, and have denser connections between them, as confirmed by examining the block structures within the top-left $46\times 46$ corner of both Figs. \ref{fig:protein} (a) and (b).
 

\subsection{NIPS234 Coauthor Network}
\vspace{-1mm}

\begin{figure}[!t]
\begin{center}
\includegraphics[width=56mm]{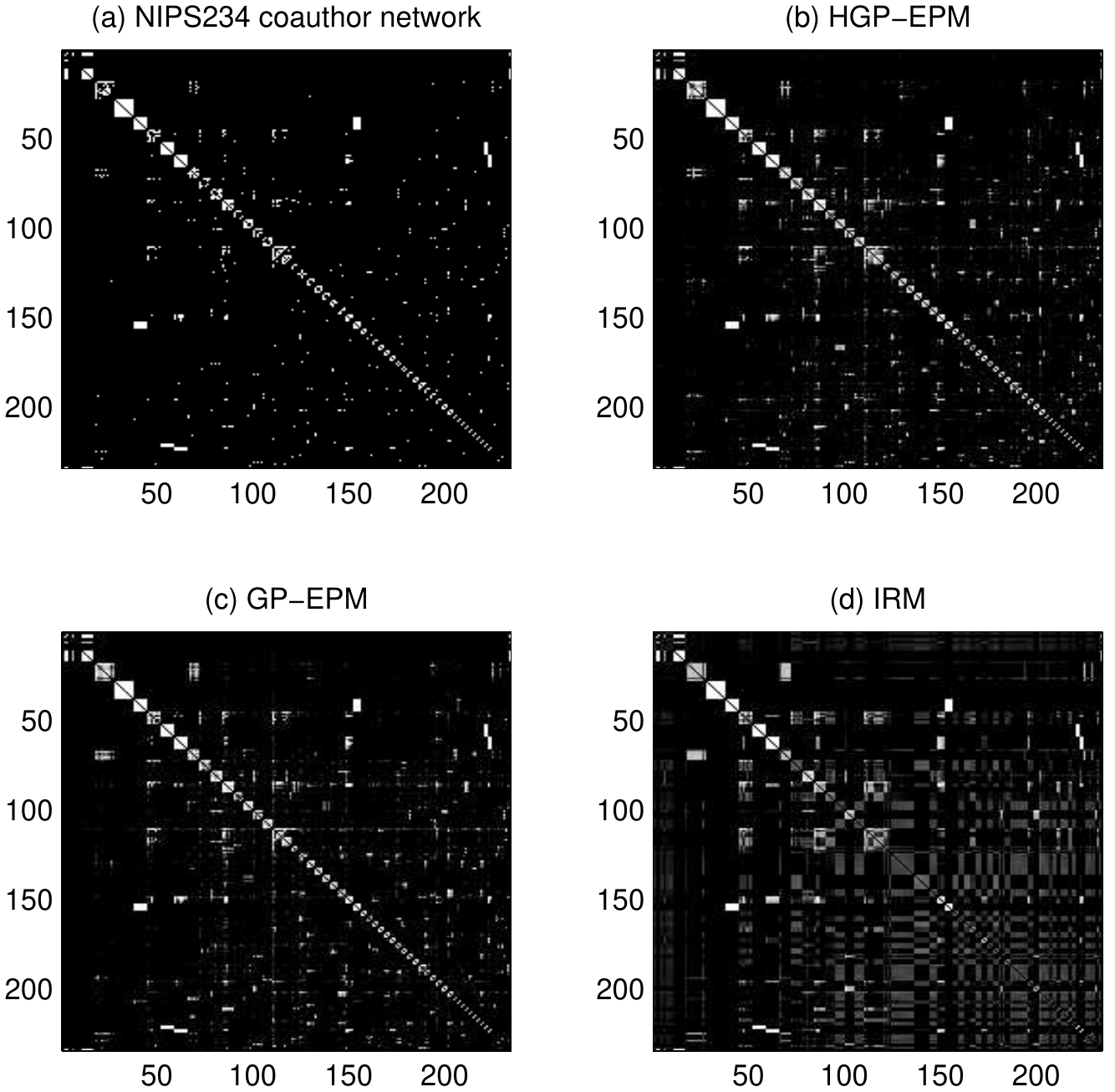}
\end{center}
\vspace{-4.9mm}
\caption{\small\label{fig:NIPS234}
Comparison of three models on estimating the link probabilities for the NIPS234 coauthor network.  (a)-(d) Analogous plots to Figures \ref{fig:protein} (a)-(d).
\vspace{-4.0mm}
}
\end{figure}

We consider the small-scale NIPS234 network consists of the top 234 authors in NIPS 1-17 conferences\footnote{\href{http://chechiklab.biu.ac.il/~gal/data.html}{http://chechiklab.biu.ac.il/$\sim$gal/data.html}} in terms of the number of publications, as studied in \citet{miller2009nonparametric}. There are 598 edges. As shown in Tab. \ref{tab:NIPS234}, the GP-EPM and HGP-EPM have the best overall performance, followed by the IRM. Comparing with the simpler GP-EPM, the extra flexibility  to model stochastic equivalence does not bring the HGP-EPM additional advantages  on this dataset, which is not surprising as  Fig. \ref{fig:NIPS234} suggests that this coauthor network mainly exhibits homophily. Note that the IRM performs well measured by the AUC-ROC, but its AUC-PR is clearly worse than those of the EPMs. This may again be explained by its overly smoothed cartoonish estimation that overlooks small communities, as clearly shown in Fig. \ref{fig:NIPS234} (d).


\begin{table}[h]\vspace{-2.5mm}\small
\caption{\small Comparison of six algorithms on predicting missing edges of the NIPS234 coauthor network.
 The Eigenmodel achieves its best performance at $K=10$.   
} 
\begin{center}
\begin{tabular}{|c|c|c|c|c|}
\hline
Model & AUC-ROC & AUC-PR\\\hline
IRM &   \textbf{0.9476} $\pm$ 0.0114&  0.6677 $\pm$ 0.0201   \\\hline
Eigenmodel &  0.9269  $\pm$ 0.0177 &  0.6784 $\pm$ 0.0364  \\\hline
ILA & 0.9171 $\pm$ 0.0222 &  0.6793 $\pm$ 0.0295 \\\hline
AGM &0.8906 $\pm$ 0.0164 &  0.5842 $\pm$ 0.0357\\\hline
GP-EPM &\textbf{0.9501} $\pm$ 0.0123 &  \textbf{0.7415} $\pm$ 0.0319\\\hline
HGP-EPM & \textbf{0.9469} $\pm$ 0.0163 &  \textbf{0.7289} $\pm$ 0.0540 \\\hline
\end{tabular}
\end{center}
\label{tab:NIPS234}
\vspace{-4.mm}
\end{table}

 \subsection{Yeast and NIPS12 Networks}
 We also consider the Yeast\footnote{\href{http://vlado.fmf.uni-lj.si/pub/networks/data/bio/Yeast/Yeast.htm}{\tiny http://vlado.fmf.uni-lj.si/pub/networks/data/bio/Yeast/Yeast.htm}} protein interaction network of \citet{yeastdatset},  with 2361 nodes and 6646 non-self edges, and the NIPS12 coauthor network\footnote{\href{http://www.cs.nyu.edu/~roweis/data.html}{http://www.cs.nyu.edu/$\sim$roweis/data.html}} that includes all the 2037 authors in NIPS papers vols 0-12, with 3134 edges. These two median-size networks are already too large for the Eigenmodel and ILA to produce reasonable results given our 
  computational resources.  
The results in Tabs. \ref{tab:Yeast} and \ref{tab:NIPS12} show that the HGP-EPM performs the best on the Yeast protein-protein interaction network, which is found to clearly exhibit  stochastic equivalence by examining the plots corresponding to the ones in Figs. \ref{fig:protein} and \ref{fig:NIPS234}  (not shown for brevity), and the HGP-EPM and GP-EPM both perform well on the NIPS12 coauthor network, which is found to mainly exhibit homophily by examining related plots (not shown for brevity).

\begin{figure}[!tb]
\begin{center}
\includegraphics[width=62mm]{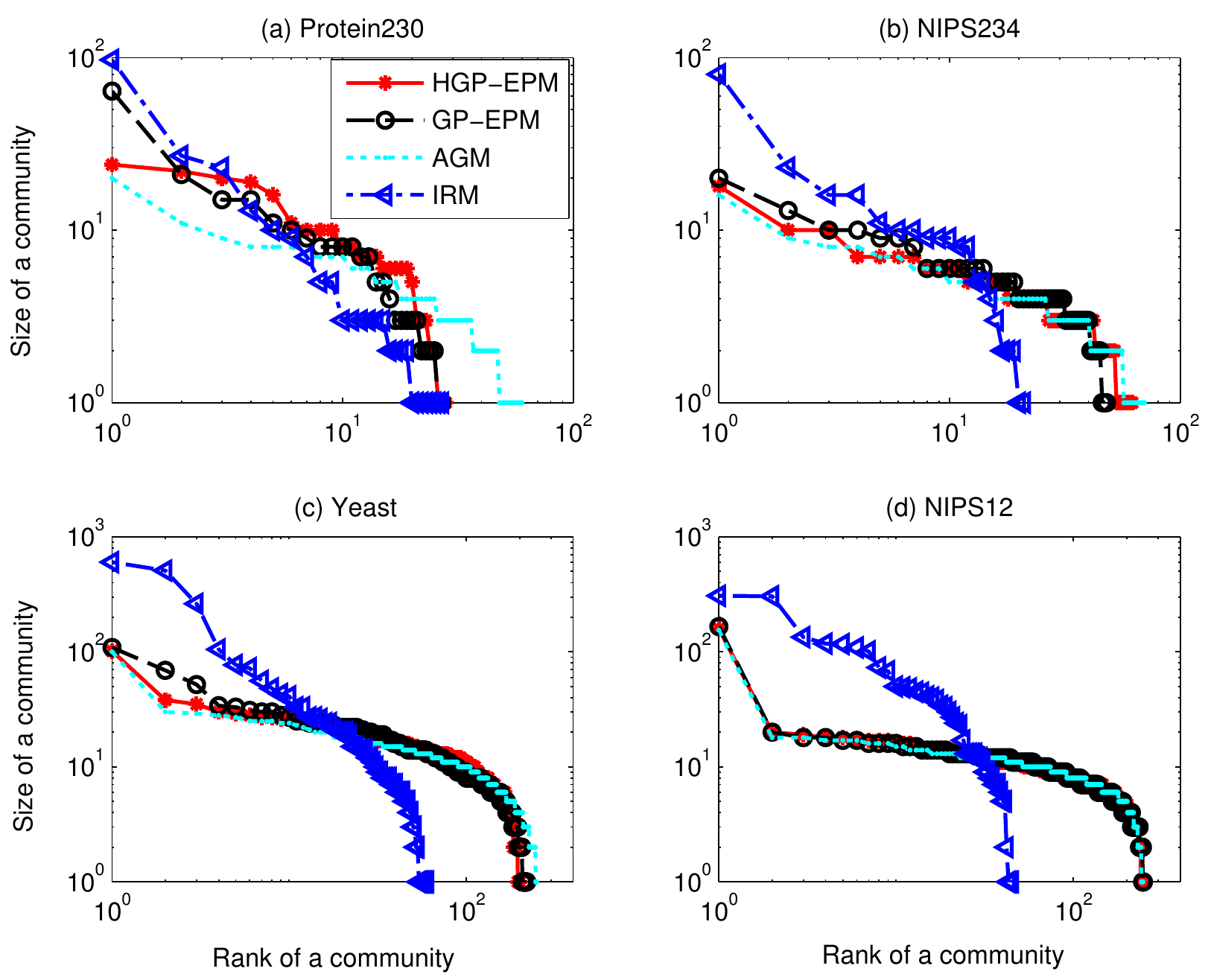}
\end{center}
\vspace{-5.0mm}
\caption{\small\label{fig:rank_size}
A community' size  and its rank.   
}\vspace{-3.5mm}
\end{figure}

As discussed before, the HGP-EPM, GP-EPM, AGM and IRM can all be used to assign nodes to disjoint communities. In Fig. \ref{fig:rank_size} we plot the size of an inferred latent community as a function of its rank (smaller ranks indicate larger sizes) on the log-10 scale,  for  the four scalable algorithms on the four tested real networks. It is clear that in contrast to the other three latent factor models, the IRM, a latent class model, infers a smaller number of communities, with more larger-size and fewer smaller-size ones. Examining the details we find that the IRM tends to place all the low-degree nodes into one or several large-size communities, whereas the other models are able to better preserve fine details involving small-size communities.

  \begin{table}[!t]\vspace{-2mm}\small
\caption{\small Comparison of four algorithms on predicting missing edges of the Yeast  protein interaction  network.  } 
\begin{center}
\begin{tabular}{|c|c|c|c|c|}
\hline
Model & AUC-ROC & AUC-PR\\\hline
IRM &   0.9093 $\pm$ 0.0059 &  0.1878 $\pm$ 0.0142   \\\hline
AGM &0.9009 $\pm$ 0.0025 &  0.1225 $\pm$ 0.0129\\\hline
GP-EPM &0.9331 $\pm$ 0.0014 &  \textbf{0.2486} $\pm$ 0.0149\\\hline
HGP-EPM & \textbf{0.9367} $\pm$ 0.0012 &  \textbf{0.2628} $\pm$ 0.0184 \\\hline
\end{tabular}\label{tab:Yeast}

\vspace{-1.0mm}

\caption{\small Comparison of four algorithms on predicting missing edges of the NIPS12 coauthor network.  
} 
\begin{tabular}{|c|c|c|c|c|}
\hline
Model & AUC-ROC & AUC-PR\\\hline
IRM &   0.9427 $\pm$ 0.0121&  0.2066 $\pm$ 0.0331   \\\hline
AGM &0.9328 $\pm$ 0.0049 &  0.2350 $\pm$ 0.0177\\\hline
GP-EPM &\textbf{0.9768} $\pm$ 0.0079 &  \textbf{0.4705} $\pm$ 0.0362\\\hline
HGP-EPM & \textbf{0.9762} $\pm$ 0.0081 &  \textbf{0.4493} $\pm$ 0.0229 \\\hline
\end{tabular}\label{tab:NIPS12}
\end{center}
\vspace{-5.7mm}
\end{table} 
    
 We mention that  the HGP-EPM, GP-EPM and AGM have 
 $O(N\bar{d} + NK)$ computation, whereas the Eigenmodel and ILA have at least 
$O(N^2+NK)$ computation, where $K$ is the number of latent features. 
With unoptimized Matlab on a 2.7 GHz CPU, for 1000 MCMC iterations, the HGP-EPM (GP-EPM) takes about 80 (20) seconds on Protein230, about 85 (28) seconds on NIPS234, about 50 (18) minutes on Yeast, and about 32 (12) minutes  on NIPS12. The Eigenmodel with $K=25$ takes about 200 seconds on NIPS234 to run 1000 MCMC iterations.
For the ILA on NIPS234, we considered 1000 MCMC iterations that took over 18 hours to run; for Protein230, the ILA inferred about two times  more features as it did on NIPS234, and we considered  500 MCMC iterations that took over 21 hours to run. 

  \vspace{-1.mm}
\section{CONCLUSIONS}
  \vspace{-1.mm}

To model unweighted undirected relational networks characterized by both homophily and stochastic equivalence, we propose a hierarchical gamma process edge partition model (EPM) that supports an infinite number of communities and an infinite square rate matrix to describe community-community interactions. 
The EPM exploits a Bernoulli-Poisson link to assign a latent count to each binary edge, and further partitions that count according to the edge's affiliations with all pairs of communities, which naturally leads to the partition of each node into overlapping communities.
We also provide a simpler gamma process EPM  that omits  inter-community interactions, which is found to perform well on assortative networks. Efficient MCMC inference with closed-form update equations is provided. Experimental results on four real networks illustrate the EPMs' working mechanisms and properties, as well as their state-of-the-art performance and interpretable latent representations. 
While previous  latent feature relational models   
and their nonparametric Bayesian versions are often not scalable, 
our infinite EPMs  are readily scalable to networks with thousands of nodes. It would be interesting to investigate strategies to make  them scalable to relational networks with millions of nodes and edges. 
 
 \nocite{BNBP_PFA_AISTATS2012}

\newpage

\newpage


\bibliographystyle{plainnat}
\bibliography{References102014}
 \newpage
  
  \newpage
  \appendix

\textbf{Infinite
Edge Partition Models for Overlapping\\ Community Detection and Link Prediction: Appendix}

\section{Proof for Lemma 1}
Using the law of total expectation, we have $$
\E\left[\sum_{k_1}\sum_{k_2} \lambda_{k_1 k_2} \right]  
=\frac 1 \beta\E\left[{\xi G(\Omega)}+{ [G(\Omega)]^2}-{\sum_k r^2_{k}}\right]
.$$ Using Campbell's theorem \citep{PoissonP}, we have $$\E\left[\sum_{k} r_k^2\right]=\int_{\Omega} \int_{0}^\infty r^2 r^{-1} e^{-c_0 r} dr G_0(d\omega) = \frac{\gamma_0}{c_0^2}.$$
The proof is completed by further  using $\E[G(\Omega)]=\gamma_0/c_0$ and $\E[G^2(\Omega)]=\gamma^2_0/c_0^2+\gamma_0/c_0^2$.
\qed

\section{MCMC Inference for HGP-EPM}
\textit{\textbf{Sample $m_{ij}$.}}  As  in Section \ref{sec:PoBer}, 
we sample 
a latent count for each $b_{ij}$ as
\begin{align}
(m_{ij}|-)\sim b_{ij}\mbox{Po}_+\left(\sum_{k_1=1}^K\sum_{k_2=1}^K \phi_{ik_1}\lambda_{k_1 k_2}\phi_{jk_2}\right).
\label{eq:sample_mij}
\end{align}
\textit{\textbf{Sample $m_{ik_1k_2j}$.}}
Using the relationships between the Poisson and multinomial distributions, similar to the derivation in 
\citet{BNBP_PFA_AISTATS2012}, we partition the latent count $m_{ij}$ as
\beq
(\{m_{ik_1k_2j}\}|-) \sim\mbox{Mult}\bigg(m_{ij}; \frac{\{\phi_{ik_1}\lambda_{k_1 k_2}\phi_{jk_2}\} }{\sum_{k_1}\sum_{k_2} \phi_{ik_1}\lambda_{k_1 k_2}\phi_{jk_2}}\bigg).
\eeq
Note that in each MCMC iteration we store $m_{i k \cdotv \cdotv}$ and $m_{\cdotv k_1 k_2 \cdotv}$ but not  necessarily  $m_{ik_1k_2j}$ in the memory.\\
\textit{\textbf{Sample $a_i$.}} 
Using (\ref{eq:a_i_likelihood}) and the data augmentation technique developed in \citet{MingyuanNIPS2012,NBP2012} for the negative binomial distribution, 
we sample $a_i$ as
\begin{align}
 &(\ell_{ik}|-) \sim \sum_{t=1}^{m_{i k \cdotv \cdotv} } 
 \mbox{Ber}\left(\frac{a_i}{a_i+t-1} \right),\notag\\ 
 &(a_i|-) \sim \mbox{Gam}\bigg(e_0+\sum_{k}\ell_{ik}, \frac{1}{f_0-\sum_{k}\ln(1-p'_{ik})}\bigg),
  \end{align}
  where with a slight abuse of notation, but for added conciseness, we use $x\sim\sum_{t=1}^m \mbox{Ber}[a/(a+t)]$ to represent  $x=\sum_{t=1}^m u_t,~u_t\sim\mbox{Ber}[{a}/({a+t})]$.\\
\textit{\textbf{Sample $\phi_{ik}$.}}  
Using (\ref{eq:phi_ik_likelihood}) and the gamma-Poisson conjugacy, 
we have
\begin{align}
(\phi_{ik}|-)\sim \mbox{Gam}\big[a_i + m_{i k \cdotv \cdotv} ,{1}/({c_i+ \omega_{ik})} 
\big]. 
\end{align}
\textit{\textbf{Sample $r_k$.}}   Similar to the inference of $a_i$, using (\ref{eq:r_k_likelihood}), 
we sample $r_k$ as
 \begin{align}
&(l_{k k_2}|-) \sim\hspace{-1mm} \sum_{t=1}^{ m_{\cdotv k k_2  \cdotv} }\label{eq:sample_lk1k2} 
\mbox{Ber}\left(\frac{ r_{k}\xi^{\delta_{kk_2}} (r_{k_2})^{1-\delta_{kk_2}}}  { r_{k}\xi^{\delta_{kk_2}} (r_{k_2})^{1-\delta_{kk_2}}   +t-1} \right),\notag\\
 & (r_k|-) \sim\mbox{Gam}\bigg[\frac{\gamma_0}{K}+ \sum_{k_2} l_{k k_2}, \notag\\
 &~~~~~~~~~~~~\frac{1}{c_0 
 -\sum_{k_2} \xi^{\delta_{kk_2}} (r_{k_2})^{1-\delta_{kk_2}}\ln\left(1-\tilde p_{k k_2}\right)} \bigg].
 \end{align}
\textit{\textbf{Sample $\xi$.}}   
We resample the auxiliary variables $l_{k k}$ using the updated $r_k$ 
and then  sample $\xi$ as
 \beq
 (\xi|-) \sim\mbox{Gam}\bigg[e_0+\sum_{k}  l_{k k}, \frac{1}{f_0 
 -\sum_{k}  r_{k}\ln\left(1-\tilde p_{k k}\right)} \bigg].
 \eeq
\textit{\textbf{Sample $\lambda_{k_1k_2}$.}}   Using (\ref{eq:lambda_likelihood}) and the gamma-Poisson conjugacy, we have
\begin{align}
(\lambda_{k_1 k_2}|-)\sim &~\mbox{Gam}\Big[ r_{k_1} \xi^{\delta_{k_1k_2}} (r_{k_2})^{1-\delta_{k_1k_2}}  +m_{\cdotv k_1 k_2 \cdotv},\notag\\
&~~~~~~~~~~~~~~~~~~~~~~~~~~~~~~~~~~ {1}/({\beta+ \theta_{k_1k_2})
} 
\Big].
%
\end{align}
\textit{\textbf{Sample $\beta$, $c_i$ and $c_0$.}} They can be sampled from gamma distributions using the conjugacy between gamma distributions, omitted here for brevity.   \\
  \textit{\textbf{Sample $\gamma_0$.}} As show in Lemma \ref{lem:E_lambda}, the mass parameter $\gamma_0$ plays an important role in determining the total sum of the infinite rate matrix $\{\lambda_{k_1k_2}\}$. Our experiments show that it could be used as a tuning parameter to impose one's prior preference
  on the number of active communities to be discovered. 
   In this paper, we impose a gamma prior as $\gamma_0\sim\mbox{Gam}(1,1)$ to let the data infer the posterior of $\gamma_0$.
  We employ an independence chain Metropolis-Hastings algorithm to sample $\gamma_0$, with 
the proposal distribution constructed as
  \beq
  Q(\gamma^\ast_0)=\mbox{Gam}\bigg(1+\sum_{k} \tilde{l}_k, \frac{1}{1 -\frac{1}{K}\sum_{k} \ln(1-\tilde{\tilde{p}}_k)}\bigg),
  \label{eq:Q} \vspace{-.5mm}
  \eeq
where
   $
   (\tilde{l}_k|-) \sim\mbox{CRT}\left( \sum_{k_2} l_{kk_2}, {\gamma_0}/{K}\right)\notag
   $
   and
  $$ \tilde{\tilde{p}}_k
   :=\frac{-\sum_{k_2} \xi^{\delta_{kk_2}} (r_{k_2})^{1-\delta_{kk_2}}\ln\left(1-\tilde p_{k k_2}\right)}{c_0-\sum_{k_2} \xi^{\delta_{kk_2}} (r_{k_2})^{1-\delta_{kk_2}}\ln\left(1-\tilde p_{k k_2}\right)}.$$
   We accept $\gamma_0^\ast$ with probability $\min\{1,\pi\}$, where $\pi$ is
   \beq
   \hspace{-1mm}\frac{\prod_{k=1}^K\hspace{-1mm} \mbox{Gam}(r_k;\gamma^\ast_0/K,1/c_0)\mbox{Gam}(\gamma^\ast_0;1,1)Q(\gamma_0)}{\prod_{k=1}^K\hspace{-1mm} \mbox{Gam}(r_k;\gamma_0/K,1/c_0)\mbox{Gam}(\gamma_0;1,1)Q(\gamma^\ast_0)},\notag
   \eeq
   which is usually greater than 50\% for the networks considered in this paper.

Each iteration of the MCMC for the HGP-EPM proceeds from (\ref{eq:sample_mij}) to (\ref{eq:Q}). 
  \vspace{-1mm}
  
  \section{
Gamma Process 
EPM
}
The gamma process EPM differs from the HGP-EPM in that it omits inter-community interactions, which leads to a simpler hierarchical model and faster computation at the expense of reduced ability to model stochastic equivalence. It is found to have good performance on assortative networks but not necessarily on disassortative ones. 

\subsection{Hierarchical Model}
The (truncated) gamma process EPM is expressed as
\begin{align}
&b_{ij} = \mathbf{1}(m_{ij}\ge 1),\notag\\
&m_{ij}=\sum_{k=1}^K  m_{ijk},~m_{ijk} \sim\mbox{Po}\left( r_k \phi_{ik}\phi_{jk}\right),\notag\\
&\phi_{ik}\sim\mbox{Gam}(a_i,1/c_{i}),~a_i\sim\mbox{Gam}(e_0,1/f_0),\notag\\
&r_k\sim\mbox{Gam}(\gamma_0/K,1/c_0),~ \gamma_0\sim\mbox{Gam}(e_1,1/f_1), 
\label{eq:BPCM}.
\end{align}
where the $\mbox{Gam}(1,1)$ prior is also imposed on  $c_0$ and $c_i$.
As $K\rightarrow\infty$, we recover the (exact) gamma process with a finite and continuous base measure. We usually set $K$ to be large enough to ensure a good approximation to the truly infinite model.

Note that
if we marginalize out  both $m_{ij}$ and $m_{ijk}$, then we have
\beq
b_{ij}
\sim\mbox{Bernoulli}\left[1 - \prod_{k=1}^K\exp\left(- r_k\phi_{ik}\phi_{jk} \right)\right].\notag
\eeq

\subsection{Gibbs Sampling}

Let the latent counts $m_{i\cdotv k}$ and $m_{\cdotv \cdotv k}$ be defined as
\begin{align}
&m_{i\cdotv k} := \sum_{j=i+1}^N m_{i jk} + \sum_{j=1}^{i-1} m_{jik},\notag\\
&m_{\cdotv \cdotv k}:= \sum_{i=1}^N\sum_{j=i+1}^N m_{i  j k } =\frac{1}{2}\sum_{i=1}^N  m_{i\cdotv k}\,\, .\notag
\end{align} Using the Poisson additive property, we have
\begin{align}
&m_{i\cdotv k}  \sim\mbox{Po}\bigg( r_k \phi_{ik} \sum_{j\neq i} \phi_{jk} \bigg)\label{eq:phi_ik_likelihood1},\\
&m_{\cdotv \cdotv k} \sim\mbox{Po}\left(r_k\frac{\sum_i\sum_{j\neq  i}\phi_{ik}\phi_{jk} }{2}\right )\label{eq:lambda_likelihood1}.
\end{align}

Marginalizing out $\phi_{ik}$ from (\ref{eq:phi_ik_likelihood1}), we have
 \begin{align}
 m_{i k \cdotv \cdotv}   \sim \mbox{NB}\left( a_i, p_{ik}'\right),
 \label{eq:a_i_likelihood1}
 \end{align}
 where
  $$p_{ik}':=\frac{ r_k\sum_{j\neq i} \phi_{jk} }{c_i+r_k\sum_{j\neq i} \phi_{jk}  } .$$
Marginalizing out $r_k$ from (\ref{eq:lambda_likelihood1}), we have 
\begin{align} m_{\cdotv \cdotv k }  \sim\mbox{NB}\left( \gamma_0/K,~\tilde p_{k}\right )\label{eq:r_k_likelihood1},
\end{align}
where 
$$
\tilde p_{k} := \frac{\sum_i\sum_{j\neq  i}\phi_{ik}\phi_{jk}}{2c_0+\sum_i\sum_{j\neq  i}\phi_{ik}\phi_{jk}}.
$$

Similar to the inference techniques used in Appendix B, one may exploit (\ref{eq:phi_ik_likelihood1})-(\ref{eq:r_k_likelihood1}) to derive 
closed-form Gibbs sampling update equations for all model parameters, omitted here for brevity.

 \section{Gamma Process AGM}  
   
 Closely related to the gamma process EPM, the hierarchical model for 
the (truncated) gamma process AGM can be expressed as
\begin{align}
&b_{ij} = \mathbf{1}(m_{ij}\ge 1),\notag\\
&m_{ij}= u_{ij} +\sum_{k=1}^K  m_{ijk},~m_{ijk} \sim\mbox{Po}\left( r_k \phi_{ik}\phi_{jk}\right),\notag\\
&u_{ij}\sim\mbox{Po}(\epsilon),~\epsilon\sim\mbox{Gam}(a_0,1/b_0),\notag\\
&\phi_{ik}\sim\mbox{Ber}(\pi_i),~\pi_i\sim\mbox{Beta}(a_1,b_1),\notag\\
&r_k\sim\mbox{Gam}(\gamma_0/K,1/c_0),~ \gamma_0\sim\mbox{Gam}(e_1,1/f_1). 
\end{align} 

We sample $r_k$, $\gamma_0$ and $c_0$ in the same way we sample them in the gamma process EPM. To sample $\phi_{ik}$, one may use (\ref{eq:phi_ik_likelihood1}) as the likelihood, under which $\phi_{ik}$ is equal to one a.s. if $m_{i\cdotv k}>0$ and is drawn from a Bernoulli distribution if $m_{i\cdotv k}=0$.
Gibbs sampling update equations for the other model parameters can be conviniently derived by exploiting conditional conjugacies, omitted here for brevity. 

\end{document}